\pdfoutput=1
\documentclass[sigconf]{acmart}
\AtBeginDocument{%
  \providecommand\BibTeX{{%
    \normalfont B\kern-0.5em{\scshape i\kern-0.25em b}\kern-0.8em\TeX}}}

\acmConference[Conference acronym 'XX]{}{2024}{}

\usepackage{algorithm}
\usepackage{algpseudocode}
\usepackage{multirow}
\usepackage{cleveref}
\usepackage{subcaption}
\usepackage{lipsum}
\usepackage{dcolumn}

\begin{document}

\title{ProbSAINT: Probabilistic Tabular Regression for Used Car Pricing}

\author{Kiran Madhusudhanan}
\email{madhusudhanan@ismll.uni-hildesheim.de}
\orcid{0000-0001-6356-8646}
\affiliation{%
\institution{Information Systems and
Machine Learning Lab \\
VWFS Data Analytics Research Center \\
  University Of Hildesheim}
  \city{Hildesheim}
  \country{Germany}
}

\author{Gunnar Behrens}
\affiliation{%
  \institution{Data Analytics \& AI Engineering \\
  Volkswagen Financial Services AG}
  \city{Braunschweig}
  \country{Germany}
}

\author{Maximilian Stubbemann}
\email{stubbemann@ismll.de}
\orcid{0000-0003-1579-1151}
\affiliation{%
\institution{Information Systems and
Machine Learning Lab  \\
VWFS Data Analytics Research Center \\
  University Of Hildesheim}
  \city{Hildesheim}
  \country{Germany}
}

\author{Lars Schmidt-Thieme}
\email{schmidt-thieme@ismll.uni-hildesheim.de}
\orcid{0000-0001-5729-6023}
\affiliation{%
\institution{Information Systems and
Machine Learning Lab\\
VWFS Data Analytics Research Center\\
  University Of Hildesheim}
  \city{Hildesheim}
  \country{Germany}
}

\renewcommand{\shortauthors}{Madhusudhanan, et al.}

\begin{abstract}
 
Used car pricing is a critical aspect of the automotive industry, influenced by
many economic factors and market dynamics. With the recent surge in online
marketplaces and increased demand for used cars, accurate pricing would benefit
both buyers and sellers by ensuring fair transactions. However, the transition
towards automated pricing algorithms using machine learning necessitates the 
comprehension of model uncertainties, specifically the ability to flag
predictions that the model is unsure about. Although recent literature proposes
the use of boosting algorithms or nearest neighbor-based approaches for swift
and precise price predictions, encapsulating model uncertainties with such
algorithms presents a complex challenge. We introduce \texttt{ProbSAINT}, a
model that offers a principled approach for uncertainty quantification of its
price predictions, along with accurate point predictions that are
comparable to state-of-the-art boosting techniques. Furthermore, acknowledging
that the business prefers pricing used cars based on the number of days the
vehicle was listed for sale, we show how ProbSAINT can be used as a dynamic forecasting
model for predicting price probabilities for different expected offer duration. Our
experiments further indicate that ProbSAINT is especially accurate on
instances where it is highly certain. This proves the applicability of its
probabilistic predictions in real-world scenarios where trustworthiness is crucial.

\end{abstract}


\maketitle

\section{Introduction}

The Used car market is a vast and complex ecosystem with millions of 
vehicles traded each year. 
For instance, the German used car market was valued at 113.2 Billion USD in 2021 and is expected to reach a value of 171.03 Billion USD by 2027, with a compound annual growth rate (CAGR) of 7.12\% during the forecasted period \cite{Report}. Consequently, there is a need for a scalable automated pricing system to accommodate this demand.
This increase in volume is accompanied by considerable volatility in the used car market. For example, during the COVID-19 pandemic, supply-chain challenges and shortage of chips and other electronic components, hindered the supply of both new and used cars, resulting in  inflated prices and higher profit margins for these vehicles. Thus, complex machine learning techniques capable of swiftly adapting to rapidly changing market conditions are necessary for pricing. Furthermore, to ensure the trustworthiness of model predictions in an automated pricing system, quantification of uncertainties is crucial.

Used car pricing is a complex task involving features that describe vehicle characteristics such as vehicle-model, brand, and registration year, as well as usage attributes like the damage, offer duration, and mileage. Moreover, temporal features such as initial listing date and the selling date are also crucial factors. Previous machine learning methods \cite{venkatasubbu2019used, samruddhi2020used} for used-car pricing rely on simple linear or neighborhood-based models. Consequently, they struggle to manage the complex mixture of numerical, categorical and temporal features mentioned above. Recent studies utilize more complex Gradient Boosted Decision Tree models (GBDTs) \cite{jawed2023pricing,electronics11182932} for used-car pricing. However, they fail to compare their results with state-of-the-art deep learning solutions or restrict experiments to smaller datasets. 
In contrast to the specific area of used car pricing, recent advances in tabular learning have moved in the direction of complex deep attention-based neural networks \cite{borisov2022deep}. In this work, we transfer and adapt these state-of-the-art models to the specific use case of used car pricing, resulting in pricing models which are able to handle complex feature sets while also providing enhanced prediction accuracies. Although numerous works exists on improving prediction accuracies on tabular dataset, uncertainty quantification has received less attention. This lack of study limits their applicability in industrial settings such as used car pricing, where trustworthiness is crucial. To address this issue, previous approaches have measured the uncertainty of model predictions as the standard deviation of values produced by an ensemble of models \cite{malinin2020uncertainty, lakshminarayanan2017simple}. However, these techniques offer uncertainty in a post-hoc manner across multiple models rather than training a single model to learn uncertainty from a distributional loss.

This emphasizes the necessity for state-of-the-art attention-based tabular model SAINT \cite{somepalli2021saint} to be enhanced to provide uncertainty quantification. Building on this premise, we introduce our novel approach, \texttt{ProbSAINT}. 
While SAINT was initially designed for tabular classification and does not inherently offer uncertainty quantification, we extend its capabilities to the probabilistic tabular regression setting. The proposed ProbSAINT model incorporates a probabilistic loss, specifically the log-likelihood loss, allowing the model to learn the distribution of the response variable instead of focusing solely on individual point predictions. This approach provides a principled method for modeling uncertainty without the use of post-hoc ensembles as mentioned before. Furthermore, to our knowledge, the evaluation of an attention-based tabular model such as SAINT \cite{somepalli2021saint} on large real-world regression datasets is lacking in the literature. We close this gap in the literature by demonstrating that both ProbSAINT and the underlying SAINT are proficient in tackling real-world regression tasks, such as used car pricing with a dataset comprising 2 million data points. 

In the context of used-car pricing, the \emph{offer duration} feature, indicating how long a specific car has been on the market at the time of sale prediction, holds significant importance. However, this information is typically unavailable when predicting the initial sales price for a used car to be listed on a selling platform. Often, the \emph{expected offer duration} represent a strategic decision by businesses, factoring in margins and the volume of similar used cars available. In other words, understanding the price elasticity of a given vehicle across multiple offer durations for optimal pricing is of interest. To address this, we propose the use of ProbSAINT for conditional-inference procedure named \emph{Probabilistic Dynamic Forecasting} to evaluate the price elasticity for multiple expected offer duration in the future. In section \ref{sec:probForecast}, we illustrate how ProbSAINT learns distinct price elasticities for multiple vehicles.

By offering more precise predictions alongside principled uncertainty values, our approach enhances the reliability of machine-learning-based pricing methods in the eyes of human experts. In doing so, it significantly reduces the requirement for ensuring trustworthy predictions through the implementation of extensive hand-crafted rules, which are frequently employed in real-world business contexts. Thus, ProbSAINT can facilitate a move towards fully automated used car pricing. Our specific contributions are as follows.

\begin{itemize}
    \item We propose \texttt{ProbSAINT}, to offer probabilistic outputs, thereby providing confidence values for used car pricing.
    \item We demonstrate the advantages of ProbSAINT empirically and qualitatively by utilizing appropriate distributional metrics.
\item We experimentally show that the underlying SAINT architecture serves as a
valuable deep learning alternative for large-scale used car price predicion in comparison to the
state-of-the-art boosting methods.
    \item We introduce and illustrate the utilization of ProbSAINT as a \emph{probabilistic dynamic forecasting} model with the ability to grasp market dynamics.
\end{itemize}

\section{Related Work}
Machine learning techniques have been previously employed for various price prediction settings, For example, real estate price prediction \cite{tchuente2022real}, house price estimation \cite{wang2018new}, inventory demand estimation  \cite{bajari2015machine} are few use-cases where machine learning solutions have already seen an impact. On the topic of used car price prediction \citet{venkatasubbu2019used} suggest the use of Lasso regression model for careful feature selection. The elimination of non-informative features was shown to improve the performance for used car price prediction within the Indian market. In another study, \citet{samruddhi2020used} propose a nearest neighbor approach for pricing used cars. However, the study ignore the temporal aspect of the problem and perform matching with the entire training dataset. Despite these efforts to apply machine learning for used car pricing, these methods primarily employ simple linear models. Recent studies utilize more complex Gradient Boosted Decision Tree models (GBDTs) \cite{jawed2023pricing,electronics11182932} for used-car pricing. However, they fail to compare their results with state-of-the-art deep learning solutions or restrict experiments to smaller datasets.

Advancement in the field of tabular regression is yet to be explored for the topic of used car pricing. Typically, gradient boosted tree algorithms like XGBoost \cite{Chen2016XGBoostAS}, LightGBM \cite{Ke2017LightGBMAH} or CatBoost \cite{prokhorenkova2018catboost} are used for tabular regression tasks due to their superior performance. \citet{shwartz2022tabular} compare the performance of several deep learning models with XGBoost method and conclude that efficient deep learning models for tabular data is still an active research area. Recent studies on using deep learning solutions for tabular data benefits from the use of attention mechanisms. 
\citet{somepalli2021saint} introduce SAINT, a Transformer model \cite{vaswani2017attention}, that utilizes self attention and inter-sample attention that rivals the performance of boosting tree algorithms for classification tasks. Although, SAINT was introduced for tabular classification tasks, \citet{borisov2022deep}, experiments on the effectiveness of using SAINT for tabular regression on a single dataset with 20 thousand samples. Here, the SAINT model, even though being the best deep learning model, was unable to outperform its gradient boosted counterparts. Evaluation on a single dataset does not provide qualitative understanding of the effectiveness of SAINT. A different study \cite{grinsztajn22} highlights the subpar performance of deep learning methods for tabular data. It concludes that, despite being the top-performing deep learning model, SAINT still falls short when compared to Gradient Boosted Decision Trees (GBDTs). In this study, we extend the application of the SAINT model for tabular regression tasks and assess its performance on a substantial real-world pricing dataset. Furthermore, as the uncertainty quantification of tabular data models is a relatively less explored area, we modify the SAINT framework to generate probabilistic outputs. This adaptation aims to increase the model’s reliability in an industrial environment

Uncertainty quantification is a crucial aspect for any decision-making process.
Previous works on the topic of price prediction ignores this aspect of the
problem \cite{jawed2023pricing,electronics11182932,samruddhi2020used}. Even
though boosting methods are highly competent within the tabular data regime,
they are restricted to a scalar value for regression. In order to predict a
Gaussian distribution output, the model requires at least two degrees of
freedom, for example, the mean and standard deviation of the predictions. Hence,
uncertainty quantification with boosted trees is a challenge. CatBoost
\cite{prokhorenkova2018catboost} and other boosting methods attempt to
circumvent this challenge by providing uncertainty over an ensemble of models
\cite{malinin2020uncertainty, sprangers2021probabilistic}. Another popular
approach, Monte Carlo (MC) dropout \cite{gal2016dropout} uses the dropout
\cite{Srivastava2014DropoutAS} regularization term to quantify uncertainty.
While several works have used MC dropout \cite{alarab2021illustrative} for
uncertainty quantification, it has been shown to represent incorrect posterior
for deeper neural networks \cite{foong2020expressiveness}. Above mentioned
methods are not a principled method to model uncertainty, as the model is not
trained for uncertainty quantification bur rather use ad-hoc post-processing to
quantify uncertainty. More principled techniques to model uncertainty have been
researched for deep learning models.  Normalizing flow models
\cite{papamakarios2021normalizing} are an emerging field of research to model exact
posterior distributions from given data with the help of a series of
bijective transformations. Even though normalizing flows are widely adopted
within the computer vision domain \cite{Abdelhamed_2019_ICCV} and time series
domain \cite{rasul2020multivariate}, extrapolation to tasks with a scalar value
output is less explored. Generalized linear models (GLMs)
\cite{nelder1972generalized} offer a principled approach to modelling
uncertainty. These can be easily incorporated to any model by
learning the distribution parameters of the target. Therefore, we adapt the
SAINT model to a Generalized Linear model that easily learns the distribution
parameters and provides probabilistic outputs.

\section{Problem Definition}

The problem of used car pricing involves learning from historic prices to
predict the final \emph{sales price} for an unseen used car. In addition, the task
of \emph{probabilistic used car pricing} involves calculating an expected price and a
measure of uncertainty for the prediction, or more general,
to predict the distribution of likely prices.

The abstract problem can be formalized as follows. Assume we have given 
a dataset $\mathcal{D} \subseteq \mathbb{R}^{D} \times \mathbb{R}$ of
$|\mathcal{D}| = N$ many
samples drawn from an unknown distribution $p$. The task is to find a
\emph{model} 
\begin{equation*}
    \hat{p}\colon \mathbb{R}^{D} \to \text{Dist}(\mathbb{R})
\end{equation*}
which predicts
for predictors $x\in  \mathbb{R}^D$
a distribution $\hat p(y \mid x)$ for the target $y$. Models should
minimize the expected Negative Log Likelihood (NLL):
\begin{align*}
      \ell(\hat p; p) :=
       -\mathbb{E}_{(x,y) \sim p}\log( \hat p(y \mid x))
\end{align*}

Models often choose a specific shape of target distribution in advance,
e.g., a normal distribution. To predict for every $x$ a distribution then
simplifies to predict the distribution's parameters, i.e., its mean and variance:
\begin{align*}
    \hat{m} & \colon \mathbb{R}^{D} \to \mathbb{R}^{2}
    \\ \hat{p}(y \mid x)  & :=  p_{\text{normal}}( y \mid \mu:= \hat m(x)_1,
       \sigma^2  := \hat m(x)_2 )
\end{align*}
and the Negative Log Likelihood loss (NLL) in
  $\hat m(x) = (\mu_x, \sigma_x^2)$ then is
\begin{equation}\label{eq:nll}
        \ell(\hat{m}; p ) = \mathbb{E}_{(x,y) \sim p}\left[\quad \frac{1}{2}\big(\text{log}(\text{max}(\sigma_x^2, \epsilon)) + \frac{(y - \mu_x)^2}{\text{max}(\sigma_x^2, \epsilon)}\big)\right]
\end{equation}
Here, $\epsilon >0$ is chosen as a small value
  to prevent divisions by $0$ and
  to provide numeric stability.

\section{Methodology: ProbSAINT}

\begin{figure}
    \centering
    \includegraphics[width=\columnwidth]{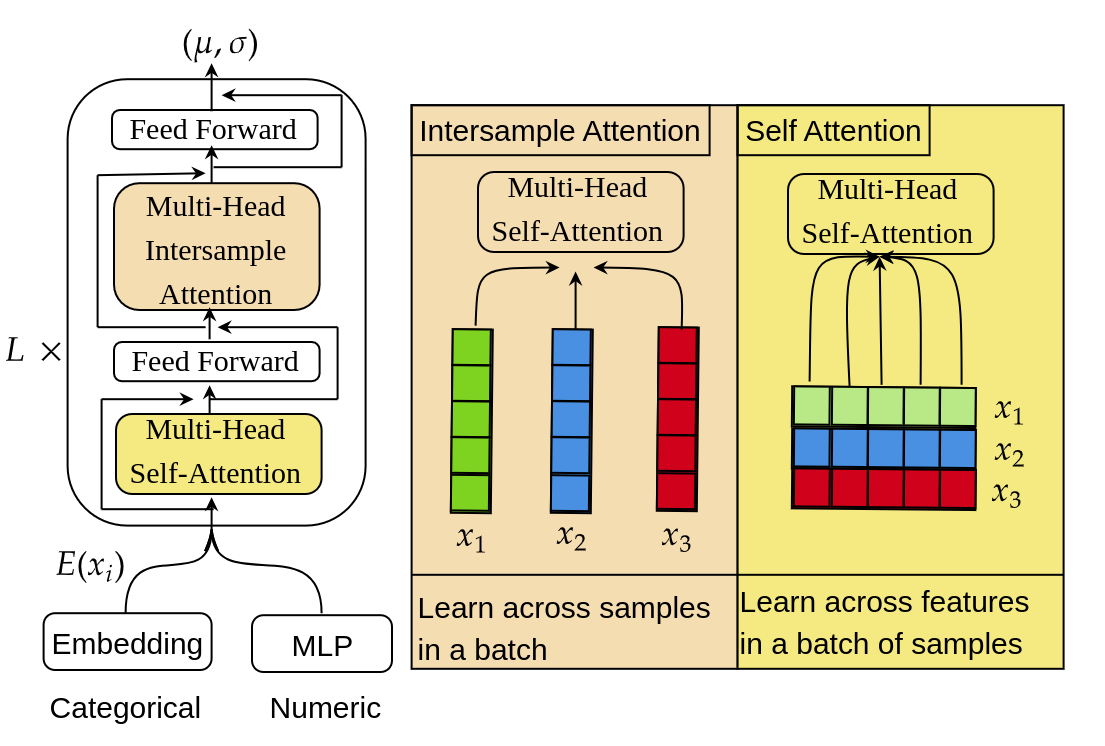}
    \caption{Architecture of proposed ProbSAINT model. Every block contains layer normalization following each attention and feed-forward layer.}
    \label{fig:ProbSAINT_archi}
\end{figure}

The recent success of deep learning can often be attributed to the expressive
power of the Transformer architecture~\cite{vaswani2017attention}. However, the direct application of the
Transformer architecture on tabular data has not witnessed significant
performance improvements similar to other domains \cite{grinsztajn22}, which has led to
attention-based tabular models that fundamentally modify the base Transformer
architecture. For example, SAINT~\cite{somepalli2021saint} adds a specific
embedding technique for numerical features and introduces inter-sample attention, 
which leads to significant performance gains in
tabular classification tasks. For more details, we refer the reader to
\citet{somepalli2021saint}. For our novel model ProbSAINT, we modify the SAINT
model to perform the task of tabular regression with uncertainty quantification.
In detail, the individual components of our model are as follows.

\subsection{Encoding}
As previously stated, the task of pricing used cars
requires a model that can effectively handle both continuous  and
categorical variables. Typically, deep learning solutions embed categorical
variables into a $d$-dimensional space, leaving numerical features unprocessed \cite{huang2020tabtransformer}.
As a result, categorical variables often dominate the embedding space. The SAINT model,
addresses this by using a separate, fully-connected network with
ReLU activation for each continuous feature. This projects the
single-dimensional numerical feature into the same $d$-dimensional vector space
where the categorical features are embedded, allowing a richer feature
representation. More specifically, our encoder $E$ maps an
input vector $x \in \mathbb{R}^D$ to a sequence of length $D$ of
$d$-dimensional vectors, i.e.,

\begin{align*}
    E \colon \mathbb{R}^D &\to (\mathbb{R}^d)^D \\
x &\mapsto (E_1(x_1), \dots, E_D(x_D)),
\end{align*}
where $E_i : \mathbb{R} \to \mathbb{R}^d$ is the encoder of the $i$-th feature
of $\mathcal{D}$.

\subsection{Self Attention and Inter-Sample Attention}
 SAINT relies on
the alternating use of self-attention and inter-sample attention. While
self-attention is a concept borrowed from the Transformer architecture \cite{vaswani2017attention},
inter-sample attention was first introduced in SAINT. Unlike the original
attention mechanism, which applies attention across different features, inter-sample
attention enables the computation of attention across different data points
within a given batch. The ProbSAINT model along with the inter-sample attention
is illustrated in \Cref{fig:ProbSAINT_archi}. The authors of the SAINT model
suggest that the inter-sample attention enables the model to borrow missing or
noisy features from other similar data points, which is beneficial in the
tabular data setting. Therefore, the SAINT model is composed of a stack of $L$
identical blocks, each of which alternates between self-attention and
inter-sample attention. 

In detail, let $z^{(1)},\dots,z^{(m)}$ be a batch of $m$ data points, where a
data point $z^{(i)}$ is a sequence of $D$ many $d$-dimensional vectors.
 We represent such datapoints via a matrix $Z^{(i)} \in \mathbb{R}^{D \times
d}$ where $Z^{(i)}_{k,l}$ is the $l$-th feature of the $k$-th
element of the sequence. We represent the batch $(Z^{(1)},\dots,Z^{(m)})$
via a tensor $Z \in \mathbb{R}^{m \times D \times d}$, where $Z_{i,k,l} 
\coloneqq Z^{(i)}_{k,l}$. 
The \emph{self-attention} layer applies the normal attention
mechanism on each datapoint over the features or columns, which can be seen as a function
\begin{equation*}
    \text{\texttt{self-attention}}_{D,d}\colon \mathbb{R}^{D \times d} \to \mathbb{R}^{D \times d}.    
\end{equation*}
A self-attention layer applies this to all datapoints of a batch individually and can be seen
as a function
\begin{equation*}
    \text{\texttt{Self-Attention}}\colon\mathbb{R}^{m \times D \times d} \to \mathbb{R}^{m \times D \times d}.    
\end{equation*}

Then, the \emph{inter-sample} layer is applied over the datapoints or rows in a batch, which works as follows: We concat all
sequences so that we have only one vector for each datapoint which can be
interpreted as a map
\begin{equation*}
    \texttt{concat}\colon \mathbb{R}^{m \times D \times d} \to \mathbb{R}^{1 \times m \times (Dd)} 
\end{equation*}
on which the normal attention mechanism is applied, for which the output is then
split back to a sequence via
\begin{equation*}
     \texttt{split}\colon \mathbb{R}^{m \times (Dd)} \to \mathbb{R}^{m \times D \times d}.
\end{equation*}

The full inter-sample attention is then given via 
\begin{align*}
    &\texttt{InterSample-Attention}\colon \mathbb{R}^{m \times D \times d} \to \mathbb{R}^{m \times D \times d} \\
    &Z \mapsto  \texttt{split}(\texttt{self-attention}_{m, Dd}(\texttt{concat}(Z))).
\end{align*}

\subsection{Distributional Output}
Our complete model is composed of an encoder and $L$ blocks of attention layers as illustrated 
in \Cref{fig:ProbSAINT_archi}. Furthermore, we incorporate a final 2-layer MLP with ReLU activation at the hidden
layer and no activation at the output, to predict means and standard
deviations. This approach differs from SAINT, where the 2-layer MLP
only produces a single output for regression target. The model is trained on the objective defined in \Cref{eq:nll}.

\begin{table}[]
    \caption{Example used-car features and corresponding data types. The ``sales price'' is our target feature.}
    \label{tbl:dataset_feat}
\resizebox{\columnwidth}{!}{%
\begin{tabular}{|ll|ll|}
\hline
\textbf{Feature}     & \textbf{Type} & \textbf{Feature}        & \textbf{Type} \\
\hline
Odometer             & Numeric       & Available Date         & Date        \\
Condition            & Categorical        & Engine Type             & Categorical        \\
Age                  & Integer       & Engine Size             & Numeric       \\
Offer Duration           & Integer       & Engine Power            & Numeric       \\
Make                 & Categorical        & Fuel Type               & Categorical        \\
Initial Registration & Date          & Transmission            & Categorical        \\
Model                & Categorical        & Euro Emissions Std & Categorical       \\ 
Model Variant        & Categorical        & Sales Date & Date       \\ 
Brand                & Categorical        & Sales Price & Numeric       \\ 
\hline
\end{tabular}
}
\end{table}

\section{Experimental Setting}
\label{sec:exp}
\subsection{Dataset} 

Given the absence of large-scale benchmark datasets for used car pricing, we train and evaluate the ProbSAINT model on externally acquired and standardized data extracted from real-world Business-to-Customer (B2C) online websites such as autoscout24, mobile.de, among others. The dataset that we derive consists of 65 features, which include 55 categorical values, 7 numerical, and 3 date features. Spanning a period from July 1, 2018, to August 20, 2022, the data comprises approximately 2 million records. Table \ref{tbl:dataset_feat} provides a summary of the key features of  the B2C dataset. We train the model on these 65 features with the aim of predicting the final sales price, also conditioned on the offer duration for a specific used car.

\subsection{Data Preprocessing}
\label{sec:data}
We convert categorical features into integer encodings using sklearn \cite{pedregosa2011scikit}. This encoding treats any new categories that may appear in the testing range as unknown. In our scenario, this could involve the introduction of a new vehicle or a different variant of an existing vehicle. We also preprocess the dataset by replacing any missing values with a placeholder value. For the ‘Date’ features in the dataset, we generate integer features such as day of the month, month of the year, and year. For the months, we create embeddings as follows.
\begin{align*}
    \text{SinMonth}(m) &= \texttt{Sin}\big(\frac{2\pi}{12}m\big), \\
    \text{CosMonth}(m) &= \texttt{Cos}\big(\frac{2\pi}{12}m\big). \\ 
\end{align*}
These periodic embeddings reflect the seasonal nature of the different months. Furthermore, we also incorporate the absolute time information by determining the number of days from a specified date to capture any trend information in the data.

\subsection{Training and Evaluation}

In our experiments, we adopt a time-wise split to divide the dataset into training, validation, and testing data. For example, if ‘2022-03-20’ is set as the beginning of the testing period, all records up to ‘2022-02-20’ form the training data. The validation data is composed of the last month’s data, spanning from ‘2022-02-20’ to ‘2022-03-20’, while the testing data includes three months of records, from ‘2022-03-20’ to ‘2022-06-20’. This time-wise split reflects real-world scenarios where the testing data lies in the future. Consequently, we create two versions of the B2C data: ‘B2C-March’, which starts the testing period on ‘2022-03-20’, and ‘B2C-June’, which starts the testing period on ‘2022-05-20’. Although retraining on the validation data could be a beneficial strategy, for the current research, we limit the model, to train only on the training split. A previous study \cite{jawed2023pricing} on a similar dataset reported minimal to no performance enhancement after retraining on the validation data.

For the task of probabilistic prediction, all the neural network models, and the NGBoost model, were trained using Negative Log Likelihood (NLL) as outlined in \Cref{eq:nll}. The CatBoost model, which offers uncertainty quantification, was trained using the ‘RMSEWithUncertainty’ loss function. Regardless of the loss function used during training, we employ NLL as a metric to assess the quality of probabilistic predictions. In addition, for the evaluation of non-probabilistic point prediction tasks, we utilize Mean Absolute Error (MAE) and Mean Absolute Percentage Error (MAPE), which are defined as follows:

\begin{align*}
    \text{MAE}(\hat{m}, p) &= \mathbb{E}_{(x,y) \sim p} |y-\mu_x|, \\
    \text{MAPE}(\hat{m}, p) &= \mathbb{E}_{(x,y) \sim p} \frac{|y-\mu_x|}{\text{max}(\epsilon, |y|)}. \\
\end{align*}




\section{Probabilistic Price Prediction}
Our results section begins with a comparison of ProbSAINT 
with other methods that offer probabilistic interpretations for 
tabular regression. We evaluate the model using the Negative Log Likelihood 
(NLL) as a probabilistic metric. Additionally, we report the deviation of the 
mean prediction ($\mu_x$) from the true selling price ($y$) using the Mean 
Absolute Percentage Error (MAPE).


The main focus of our research is to determine if the ProbSAINT model, which is trained on a distribution loss, can outperform the  state-of-the-art boosting alternatives in quantifying uncertainty. In the following section, we first present the chosen baselines for our experiments, followed by an evaluation of the results in \Cref{tab:nll}.
\paragraph{Baselines.}
CatBoost, the system currently in use, serves as a primary baseline for our research. For uncertainty quantification, CatBoost provides an out-of-the-box uncertainty quantification measure when trained using the "RMSEWithUncertainty" loss function, a model we refer to as "CatBoostUn". We describe below the core ideas for our chosen baselines.    
\label{sec:baselines}
\begin{enumerate} 
    \item MC-Dropout \cite{gal2016dropout} (ICML'16): MC-Dropout uses dropout during the testing phase as an approximate Bayesian inference in deep Gaussian processes. Here, we use the best performing SAINT model as the backbone for this method.
    \item NGBoost \cite{duan2020ngboost} (ICML'20): NGBoost is a modular framework that learns using the "Natural Gradient" descent as an alternative to stochastic gradient descent. NGBoost considers the parameters of a conditional distribution as targets for a multiparameter boosting algorithm.  
    \item CatBoostUn \cite{malinin2020uncertainty} (ICLR'21): The well known CatBoost library allows uncertainty quantification of its predicted values when trained with the loss function "RMSEWithUncertainty". The model predicts the uncertainty from an ensemble-of-trees. The study claims that ensembles of gradient boasting models were able to successfully detect anomalous inputs.   
    \item ProbMLP: In order to contrast the importance of the SAINT architecture backbone in ProbSAINT, we use a deep Multi-Layer Perceptron (MLP) model as the base architecture.
\end{enumerate}

\begin{table}[]
    \caption{Performance comparison of the proposed ProbSAINT model with relevant baselines on probabilistic metric of NLL and point prediction MAPE metric. (lower is better).}
    \resizebox{\columnwidth}{!}{%
    \begin{tabular}{lr@{}lr@{}l|r@{}lr@{}l}
    \hline
               & \multicolumn{4}{c|}{NLL}                                      & \multicolumn{4}{c}{MAPE}                                     \\
               & \multicolumn{2}{c}{B2C-March} & \multicolumn{2}{c|}{B2C-June} & \multicolumn{2}{c}{B2C-March} & \multicolumn{2}{c}{B2C-June} \\ \hline
    MC-Dropout & 22.145&$\,\pm\,$1.128        & 11.099        & $\,\pm\,$0.256        & \underline{0.058}         & \underline{$\,\pm\,$0.000}         & \underline{0.056}         & \underline{$\,\pm\,$0.000}        \\
    NGBoost    & 8.462&$\,\pm\,$0.058        & 8.228         & $\,\pm\,$0.017        & 0.059         & $\,\pm\,$0.001         & 0.055         & $\,\pm\,$0.001        \\
    CatBoost   & 8.491          &$\,\pm\,$0.003        & \underline{8.217}         & \underline{$\,\pm\,$0.001}        & 0.063         & $\,\pm\,$0.000         & 0.057         & $\,\pm\,$0.000        \\
    ProbMLP    & \underline{8.355}          &\underline{$\,\pm\,$0.076}        & 8.314         & $\,\pm\,$0.027        & 0.065         & $\,\pm\,$0.004         & 0.065         & $\,\pm\,$0.003        \\
    ProbSAINT  & \textbf{8.192}          & \textbf{$\,\pm\,$0.026}        & \textbf{8.144}         & \textbf{$\,\pm\,$0.022}        & \textbf{0.056}         & \textbf{$\,\pm\,$0.001}         & \textbf{0.053}         & \textbf{$\,\pm\,$0.001}       \\ \hline
    \end{tabular}
    }
    \label{tab:nll}
\end{table}

\paragraph{Analysis.}
Table \ref{tab:nll} presents a comparison of the proposed ProbSAINT with other
methods used for uncertainty quantification. 
As observed from Table
\ref{tab:nll}, despite extensive modifications to these GBDT methods, the NLL
loss for these methods is higher than that of the proposed ProbSAINT model on
both dataset splits. Interestingly, the high-quality
probabilistic output from ProbSAINT does not compromise point prediction
accuracy, as evidenced by the MAPE values for the corresponding models. The
second best model in terms of the MAPE value is the MC-Dropout method that uses
a SAINT backbone. The fact that the MC-Dropout model performs second best on the
MAPE values is not a surprise here, as the underlying SAINT model was trained
and tuned for the point prediction task. However, for the uncertainty
quantification task the MC-Dropout method, that uses the dropout at testing phase,
fails to capture uncertainty as shown by the NLL error rates. To summarize,
while GBDT models are reliable for quick and accurate point predictions,
ProbSAINT emerges as a far superior alternative for probabilistic forecasting.

\begin{figure}
    \centering
    \includegraphics[width=\columnwidth]{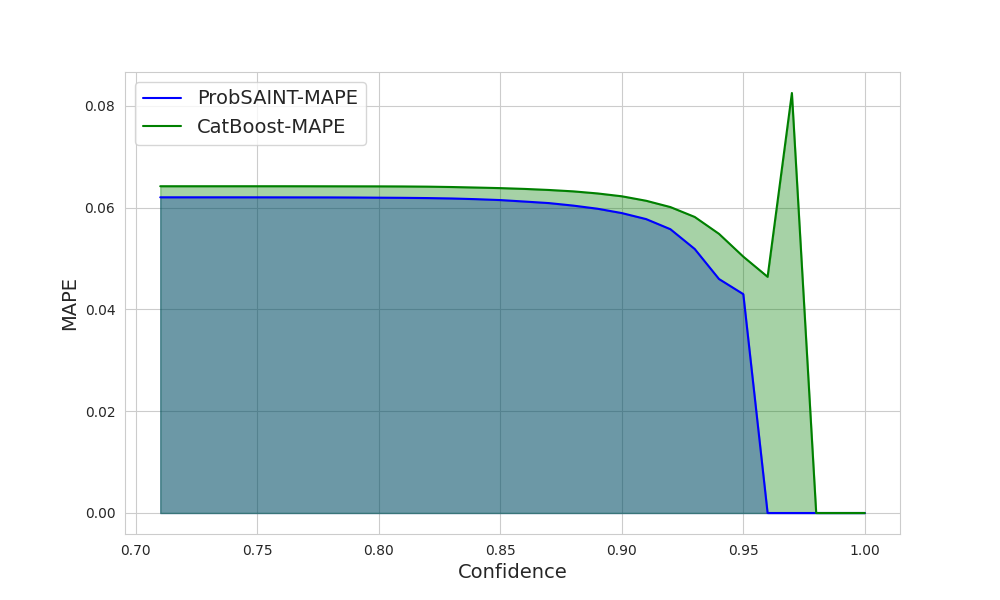}
    \caption{Qualitative comparison of ProbSAINT with that of the second best CatBoostUn baseline. The x-axis denotes the confidence score and y-axis indicates the corresponding MAPE error.}
    \label{fig:uncer_MAPE}
\end{figure}

\subsection{Probabilistic Prediction Quality at Multiple Confidence Levels}
To further evaluate the benefits of employing ProbSAINT for uncertainty
quantification, we graphically represent the confidence and the corresponding
error for ProbSAINT and compare it with that of the CatBoostUn method, as
depicted in \Cref{fig:uncer_MAPE}. In this context, we compute the
confidence scores via

\begin{align*}
    C = 1 - \frac{\sigma_x}{\mu_x},
\end{align*}
where $\mu_x, \sigma_x$ indicates the mean and standard deviation of the
prediction, respectively. As depicted in Fig. \ref{fig:uncer_MAPE}, CatBoostUn
tends to make overconfident predictions, resulting in a larger area under the
curve and outliers compared to ProbSAINT. More specifically, one can observe
that the CatBoost error increases for highly confident predictions. Furthermore,
even in the area of 0.93 to 0.95 confidence, the MAPE difference between
ProbSaint and CatBoost increases. Thus, the ability of ProbSAINT to learn the
standard deviation and mean using the NLL loss provides it with a clear
advantage over the baseline. 

\begin{figure}
    \begin{subfigure}{\columnwidth}
        \includegraphics[width=\columnwidth]{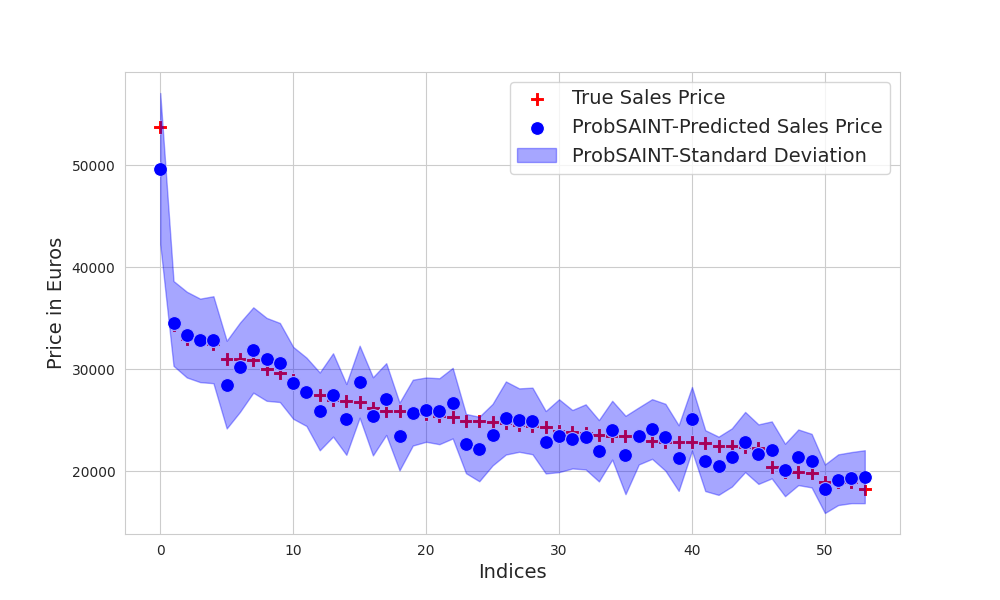}
        \caption{ProbSAINT}
        \label{fig:ProbSAINT_baseline_1}
    \end{subfigure}
        
        \bigskip
        
    \begin{subfigure}{\columnwidth}
        \includegraphics[width=\columnwidth]{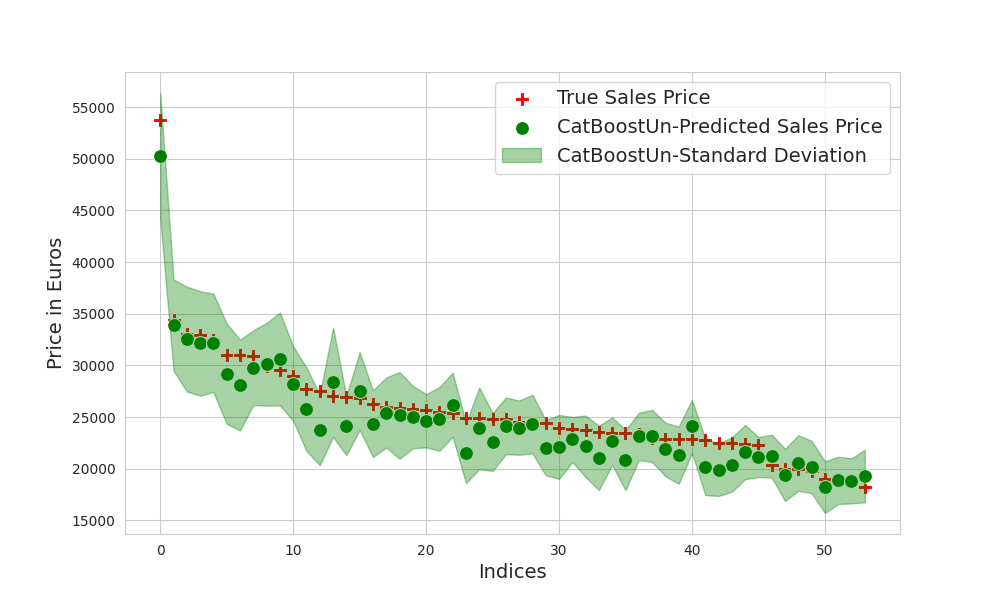}
        \caption{CatBoostUn}
        \label{fig:ProbSAINT_baseline_2}
    \end{subfigure}
    \caption{Qualitative analysis of ProbSAINT and CatBoostUn baseline. X-axis denotes the indices of different vehicles order in the decreasing order of their true selling price and Y-axis denotes the price.}
\label{fig:individual}
\end{figure}



\subsection{Individual Price Predictions}
We furthermore compare the standard deviation and mean from ProbSAINT and
CatBoostUn by plotting the difference to the true value
in~\Cref{fig:individual}. Here, the interesting observation is, that
ProbSAINT slightly overpredict prices while CatBoost slightly underprices.
However, for both methods the true value is always within the standard deviations,
which indicates that both methods produce reasonable distributions.

\subsection{Point Price Prediction}

The probabilistic prediction results clearly show the advanatage of
ProbSAINT. We now investigate wether the underlying model of ProbSAINT, namely SAINT, serves as a
potent model for tabular regression on larger datasets. To address this
auxiliary research question, we base our study on a previous survey paper
\cite{borisov2022deep}. This study compares various deep learning techniques
with GBDTs and concludes that GBDTs outperform deep learning techniques for
tabular tasks. However, the authors’ evaluation of their methods is based on a
considerably smaller dataset, comprising nearly 20 thousand records for tabular
regression, in contrast to the used-car pricing dataset, which contains 2
million rows. Consequently, we strive to answer this significantly relevant
research question using results from \Cref{tab:mae}.
\paragraph{Baselines.}
As mention before, we consider GBDT methods as a strong baseline for our analysis as it has been shown to outperform other models in recent studies \cite{borisov2022deep,grinsztajn22}. We choose additional baselines considering their superior performance on the previous study \cite{borisov2022deep}. These are as follows.
\begin{enumerate}
    \item KNN : K-Nearest Neighbor (KNN), identifies most similar data samples from the training data corresponding to the testing sample and averages the prediction across $K$ many samples.
    \item MLP: Multi-Layer Perceptrons (MLP) are deep learning solutions that employ multiple layers of fully connected layers with ReLU non-linearity in between each layer. 
    \item LightGBM \cite{Ke2017LightGBMAH} (NeurIPS'17): A powerful gradient boosting technique that speeds up the conventional Gradient Boosted Decision Tree's (GBDT), by ignoring instances with smaller gradients from the calculation of information gain and reduces the number of features by bundling mutually exclusive features.
    \item XGBoost \cite{Chen2016XGBoostAS} (KDD'16): Another powerful scalable and effective gradient boosting technique widely used to achieve state-of-the-art performance in many machine learning challenges.
    \item CatBoost \cite{prokhorenkova2018catboost} (NeurIPS'18): Gradient boosting technique that uses the target statistics to encode high fidelity categorical features into numerical features. This allows CatBoost models to be trained with almost no preprocessing steps for efficient handling and with minimum information loss. 
    \item DeepFM \cite{Guo2017DeepFMAF} (IJCAI'17): A factorization-machine based neural network initially proposed for Click-Through-Rate (CTR) prediction. The model was shown to be competitive in \cite{borisov2022deep}.
\end{enumerate}

We also experimented with additional baselines like the TabTransformer \cite{huang2020tabtransformer} and the FT-Transformer \cite{gorishniy2021revisiting} for the used car pricing task. These models encode each categorical variable with an attention layer, leading to large memory requirements to process the 2 million records with 56 categorical features in the used-car dataset. Hence, these model results could not be reported in the final evaluations. Additionally, recent work \cite{jawed2023pricing} on used-car pricing compares these models on a much smaller dataset size and report almost double the error as that of the GBDT models for point prediction.

\paragraph{Analysis.} 


From the Table \ref{tab:mae}, we note that for larger datasets, such as the
used-car pricing dataset, the SAINT model demonstrates performance on par with
the GBDT models. The SAINT model outperforms in the ‘B2C-March’ split and
exhibits comparable accuracy in the ‘B2C-June’ split for both MAPE and MAE
metrics. Additionally, the results from Table \ref{tab:nll} indicate that for
‘B2C-June’ the SAINT model is capable of similar performance as the GBDT
methods in Table \ref{tab:mae}. It is also
worth noting that the standard deviation values of the deep learning models are
larger compared to the GBDT models, suggesting a higher sensitivity to random
sources like parameter initialization. However, when compared to
\cite{borisov2022deep} and \cite{jawed2023pricing}, where deep learning models
face challenges with the regression task compared to GBDT models, we observe a
clear indication that deep learning models could close the performance gap as
the size of the dataset increases. However, the superiority of SAINT is not as
clear as the one of ProbSAINT in the probabilistic setting. This indicates that
our novel probabilistic training is a key factor for successful predictions.

\begin{table}[]
    \caption{Performance comparison of various tabular regression models on used-car pricing datasets. The models are evaluated on MAPE and MAE metrics. (lower is better).}
    \resizebox{\columnwidth}{!}{%
    \begin{tabular}{lr@{}lr@{}l|r@{}D{.}{.}{3}r@{}D{.}{.}{3}}
    \hline
             & \multicolumn{4}{c|}{MAPE}                                     & \multicolumn{4}{c}{MAE}                                      \\ 
             & \multicolumn{2}{c}{B2C-March} & \multicolumn{2}{c|}{B2C-June} & \multicolumn{2}{c}{B2C-March} & \multicolumn{2}{c}{B2C-June} \\ \hline
    KNN      & 0.175         & $\,\pm\,$0.002         & 0.173         & $\,\pm\,$0.002        & 6346.18$\,\pm\,$        & 102.68       & 6398.92$\,\pm\,$       &146.49       \\
    MLP      & 0.064         & $\,\pm\,$0.001         & 0.067         & $\,\pm\,$0.001        & 2018.22$\,\pm\,$        & 30.60         & 2133.89$\,\pm\,$       &30.66        \\
    LightGBM & 0.058         & $\,\pm\,$0.000         & \textbf{0.053}         & $\,\pm\,$0.000        & 1846.80$\,\pm\,$        & 5.76         & 1732.12$\,\pm\,$       &0.00            \\
    XGBoost  & 0.058         & $\,\pm\,$0.000         & 0.054         & $\,\pm\,$0.000        & 1847.37$\,\pm\,$       & 0.00        & 1737.77$\,\pm\,$       & 0.00            \\
    CatBoost & 0.058         & $\,\pm\,$0.000         & \textbf{0.053}         & $\,\pm\,$0.000        & 1860.81$\,\pm\,$        & 1.64         & \textbf{1713.94}$\,\pm\,$       &7.89         \\
    DeepFM   & 0.058         & $\,\pm\,$0.000         & 0.065         & $\,\pm\,$0.002        & 1865.43$\,\pm\,$        & 8.90          & 2006.16$\,\pm\,$       &10.3         \\
    SAINT    & \textbf{0.056}         & $\,\pm\,$0.001         & 0.055         & $\,\pm\,$0.002        & \textbf{1834.51$\,\pm\,$}       & 41.19        & 1782.04$\,\pm\,$      &68.07        \\ \hline
    \end{tabular}
    }
\label{tab:mae}
\end{table}

\begin{algorithm}
    \caption{Dynamical Forecasting for a specific vehicle with ProBSAINT}
    \begin{algorithmic}[1]
        \Require Trained ProbSAINT, Vehicle v with feature vector $x$
        \State Outputs = []
        \State $i$ = coordinate of $x$ which displays the offer days
        \State $O$ = [15, 45, 75, 105, 150] 
            \For {$o$ in $O$}
                \State $x_i$ = $o$ 
                \State $\mu$, $\sigma$ = ProbSAINT$(x)$
                \State Outputs.append([$\mu,\sigma$])
        \EndFor
    \end{algorithmic}
    \label{alg:dynamic}
\end{algorithm}

\section{Probabilistic Dynamic Forecasting}
\label{sec:probForecast}

The process of selling a used car involves a balancing act between the profit
margin on the sale of a used-car and the volume of used-cars sold. In certain
situations, it may be advantageous to sell the used-car as quickly as possible.
Conversely, in other scenarios, it might be more beneficial to set a higher
price for the used-car and allow it to remain listed on the platform for a
longer duration.

The data considered in \Cref{sec:exp} includes the feature \emph{offer durations}
which indicates, the duration a specific car has been on the market at the time 
of sale. While this feature is beneficial during training and testing,
it does not help in predicting the initial sales price for a new car
entering the market. Here, a reasonable procedure is to identify a
maximum of days $d_{\max}$ that the car should be on the market, allow the model to predict
the sales price with the offer durations being set to $0$ and with the offer durations
being set to $d_{max}$ and choose a sales price between the two predicted values.
This raises the question how ProbSAINT behaves with respect to varying this
feature. Therefore, in this section, we investigate how the predicted prices vary for
different models and different offer durations.

In this scenario, the task of pricing used cars can be viewed as a temporal
forecasting task, where the objective is to predict the future price of
an unknown used-car over various offer durations. However, unlike typical
forecasting tasks, we do not have time-varying information for each used-car.
Instead, we have a data point indicating the price at which the used-car was
sold. Our used-car dataset comprises 2 million such data points, representing
used-cars sold at various \emph{offer durations}. The question then is: \textbf{Can a model
trained on these 2 million records learn the general market dynamics of how the
selling price fluctuates across different offer durations for different
used-cars?}

\begin{figure}
    \centering
    \includegraphics[width=\columnwidth]{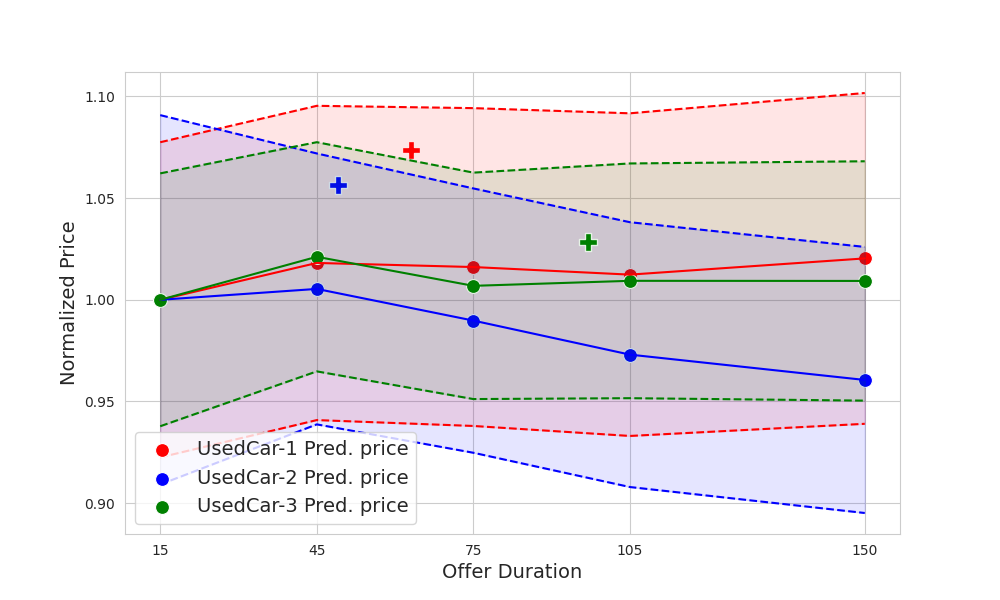}
    \caption{Probabilistic prediction of price with respect to change in offer durations. The X-axis denotes the expected offer durations for a used-car, and the Y-axis denotes the normalized price prediction from the ProbSAINT for varying offer durations. The '+' data point denotes the "true selling  price" for the "true offer durations", and the 'o' denotes predictions. Comparing three different model instances, the ProbSAINT learns varying market dynamics for the different models.}
    \label{fig:dynamic1}
\end{figure}

To answer this, we use a ProbSAINT model trained on B2C-March split,
and choose 3 different used-cars for which all features except
\emph{offer durations} are fixed. Then, we let ProbSAINT predict the output
distribution parameters $\mu$ and $\sigma$ for different amount of offer durations,
ranging from 15 to 150. This
procedure of generating dynamical forecasts with ProbSAINT is shown in Algorithm \ref{alg:dynamic}.
\Cref{fig:dynamic1} presents three distinct used cars, denoted as UsedCar-1,
UsedCar-2, and UsedCar-3, each sold at different \emph{Offer durations}. For instance,
UsedCar-1 (represented in red) was sold after 62 days on the platform, while
UsedCar-2 (in blue) and UsedCar-3 (in green) were sold after 48 and 98 days,
respectively, as indicated by the plus symbol. The y-axis is normalized by the
predicted price at 15 \emph{Offer duration} to capture the underlying trend.

\paragraph{Analysis}
 Our results provide evidence that each model learns distinct market
dynamics when it remains on the platform for an extended period. While UsedCar-1
and UsedCar-3's price predictions do not show a clear trend with respect to
increasing offer durations, ProbSAINT predicts a decreasing price trend for UsedCar-2 with
an increasing offer duration. A possible outcome from the prediction is that
the UsedCar-2 should be priced by risking a lower selling price due to the decreasing trend.
It is further notable, that for all three models
the price predicted after $45$ offer durations is higher than after $15$. Thus,
ProbSAINT indicates that keeping vehicles on the market for a short amount of
time can be beneficial in contrast to immediate sales. We are of the opinion that  
understanding of these market dynamics could be advantageous in determining the best price for used cars.

\begin{table}[]
    \caption{Hyperparameter configurations of various models used in the study.}
\label{tab:hp}
\resizebox{\columnwidth}{!}{%
\begin{tabular}{c|lll}
\hline
Model                                                                                   & \multicolumn{1}{c}{Hyperparameters} & \multicolumn{1}{c}{Type} & \multicolumn{1}{c}{Values}                      \\ \hline
\multirow{4}{*}{\begin{tabular}[c]{@{}c@{}}SAINT\\ ProbSAINT\\ MC-Dropout\end{tabular}} & Dim                                 & Categorical              & {[}32, 64, 128, 256{]}                          \\
                                                                                        & Depth                               & Categorical              & {[}1, 2, 3, 6, 12{]}                            \\
                                                                                        & Heads                               & Categorical              & {[}2, 4, 8{]}                                   \\
                                                                                        & Dropout                             & Categorical              & {[}0, 0.1, 0.2, 0.3, 0.4, 0.5, 0.6, 0.7, 0.8{]} \\ \hline
\multirow{3}{*}{MLP}                                                                    & Dim                                 & Ordinal                & (10-100)                                       \\
                                                                                        & Num. layers                         & Ordinal                & (2, 5)                                         \\
                                                                                        & Learning Rate                       & Continous                & (0.0005, 0.001)                                \\ \hline
\multirow{3}{*}{NGBoost}                                                                & Learning Rate                       & Continous                & (001, 0.3)                                     \\
                                                                                        & Base                                & Categorical              & {[}DecisionTree, GradBoost{]}               \\
                                                                                        & Mini Batch Frac                     & Continous                & (0.6, 1.0)                                     \\ \hline
\multirow{3}{*}{\begin{tabular}[c]{@{}c@{}}CatBoost\\ CatBoostUn\end{tabular}}          & Learning Rate                       & Continous                & (0,01, 0.3)                                    \\
                                                                                        & Max Depth                           & Ordinal                & (2, 12)                                        \\
                                                                                        & L2 Leaf Reg.                        & Continous                & (0.5, 30)                                      \\ \hline
DeepFM                                                                                  & Droupout                            & Continous                & (0, 0.9)                                       \\ \hline
\multirow{4}{*}{XGBoost}                                                                & Max Depth                           & Ordinal                & (2, 12)                                        \\
                                                                                        & Alpha                               & Continous                & (1e-8, 1.0)                                    \\
                                                                                        & Lambda                              & Continous                & (1e-8, 1.0)                                    \\
                                                                                        & Eta                                 & Continous                & (0.01, 0.3)                                    \\ \hline
\multirow{4}{*}{LightGBM}                                                               & Learning Rate                       & Continous                & (0,01, 0.3)                                    \\
                                                                                        & Lambda L1                           & Continous                & (1e-8, 10)                                     \\
                                                                                        & Lambda L2                           & Continous                & (1e-8, 10)                                     \\
                                                                                        & Num. Leaves                         & Ordinal                & (2, 4096)                                      \\ \hline
\multicolumn{1}{c|}{KNN}                                                                 & Neighbors                           & Ordinal                & (3, 42)                                       \\ \hline
\end{tabular}
}
\end{table}

\section{Hyperparameters}
For hyperparameter tuning each model, we use Optuna \cite{optuna_2019} with the default
Tree-structured Parzen Estimator (TPE) algorithm. The TPE algorithm is a
Sequential Model Based Optimization strategy (SMBO) that suggests the next best
trial parameters conditioned on the evaluated trials from a pre-defined
hyperparameter space. This facilitates effective hyperparameter search within an
allocated budget. For our experiments, we set the budget as 20 trials for the
Gradient Boosted Decision Tree (GBDT) models and 10 trials for the neural network
models. The trials were defined considering the fact that the neural network
models require longer training times in comparison to GBDT models. The model
specific hyperparameters are listed in \Cref{tab:hp}.

\section{Deployed System}
Our currently deployed system utilizes a CatBoost model similar, but not exactly as described in section \ref{sec:baselines} that predicts the ‘sales price’ based on the 65 features mentioned in section \ref{sec:data}. The CatBoost pricing model, deployed on AWS, is live for the German market and was subsequently expanded to include the Italian, Spanish, and Czech markets. In 2023, the CatBoost solution predicted almost 2M pricing requests, with a human expert acceptance rate of 50-60\%.

\section{Conclusion and Future Work}
We presented ProbSAINT, a novel deep neural network for probabilistic regression
on tabular data. By adapting the well-established SAINT approach and incorporating
a 2-layer MLP head, ProbSAINT is able to predict distributions instead of point predictions.
Thus, ProbSAINT is able to provide confidence values for its predictions. This is especially
required in real-world applications where trustworthiness is crucial, such as
predicting for used car prices. Our experiments on real-world industry datasets
show that ProbSAINT outperforms boosting-based state-of-the-art 
baselines for probabilistic regression tasks. Further experiments indicate that ProbSAINTs prediction are
particularly accurate when it has a high confidence, validating that ProbSAINT
computes reasonable target distributions. Finally, we demonstrate that ProbSAINT
can be used to compute prices for multiple offer durations, allowing it to
include background information like the maximal desired time in
the market.

In this study, we trained ProbSAINT using an end-to-end approach, without the 
use of auxiliary or self-supervised pre-training tasks. In future work, we plan to develop
useful pre-training tasks to leverage larger volume of additional data.
However, determining the optimal pre-training strategy for probabilistic industrial
applications such as used car pricing is an open problem. 
ProbSAINT's ability to provide reliable point- and
probabilistic outputs paves the way for future work on adapting established deep learning 
techniques like transfer learning and self-supervised representation
learning.

\bibliographystyle{ACM-Reference-Format}
\bibliography{bibliography.bib}

\end{document}